\documentclass[preprint,12pt]{elsarticle}

\usepackage{amssymb}
\usepackage{amsmath}
\usepackage{booktabs}
\usepackage{multirow} 
\usepackage{adjustbox}
\usepackage{subfig}
\usepackage{tabularx}
\usepackage{enumitem}

\journal{Expert Systems with Applications}

\begin{document}

\begin{frontmatter}



\title{Network-Based Detection of Autism Spectrum Disorder Using Sustainable and Non-invasive Salivary Biomarkers} 


\author[label1]{Janayna M. Fernandes \corref{cor1}} 
\ead{jmfernandes@ufu.br}

\author[label2]{Robinson Sabino-Silva} 
\ead{robinsonsabino@gmail.com}

\author[label1]{Murillo G. Carneiro}
\ead{mgcarneiro@ufu.br}

\cortext[cor1]{Corresponding author.}


\affiliation[label1]{organization={Faculty of Computing, Federal University of Uberlândia},country={Brazil}}
\affiliation[label2]{organization={Department of Physiology, Institute of Biomedical Sciences, Federal University of Uberlândia},country={Brazil}}

\begin{abstract}
Autism Spectrum Disorder (ASD) lacks reliable biological markers, delaying early diagnosis. Using 159 salivary samples analyzed by ATR-FTIR spectroscopy, we developed GANet, a genetic algorithm-based network optimization framework leveraging PageRank and Degree for importance-based feature characterization. GANet systematically optimizes network structure to extract meaningful patterns from high-dimensional spectral data. It achieved superior performance compared to linear discriminant analysis, support vector machines, and deep learning models, reaching 0.78 accuracy, 0.61 sensitivity, 0.90 specificity, and a 0.74 harmonic mean. These results demonstrate GANet’s potential as a robust, bio-inspired, non-invasive tool for precise ASD detection and broader spectral-based health applications.
\end{abstract}




\end{frontmatter}




\section{Introduction}\label{sec:intro}

Autism Spectrum Disorder (ASD) comprises a group of complex neurodevelopmental conditions characterized by difficulties in social interaction and communication, restricted interests, and repetitive behaviors, typically emerging during early childhood \cite{zeidan2022global,american2013diagnostic}. Global prevalence estimates indicate that ASD affects approximately 1 in 100 children worldwide, with reported rates increasing due to greater awareness and improved diagnostic protocols \cite{zeidan2022global, maenner2020prevalence}.

ASD has a multifactorial etiology involving genetic, epigenetic, and environmental factors that influence early brain development \cite{gardener2011prenatal, sandin2014familial}. Studies have identified various genetic mutations and copy number variations associated with ASD, indicating a strong heritable component \cite{gratten2016rare, sandin2017autism}. Environmental factors, including prenatal exposures, maternal immune activation, and advanced parental age, have also been implicated in increasing ASD risk \cite{lyall2017autism, bengtsson2023maternal}.

Despite global advances toward early diagnosis, many individuals with ASD continue to remain undiagnosed until adolescence or adulthood, particularly in under-resourced countries \cite{malik2022tackling, shaw2020early}. Diagnosing ASD typically involves comprehensive behavioral assessments conducted by multidisciplinary teams using standardized tools, such as the Autism Diagnostic Observation Schedule and the Autism Diagnostic Interview-Revised \cite{lord2012autism}. However, behavioral assessments can be subjective, time-consuming, and may delay diagnosis, particularly in low-resource settings, leading to missed opportunities for early intervention \cite{malik2022tackling, shaw2020early}. Early diagnosis is crucial, as targeted behavioral therapies initiated during the critical developmental window can significantly improve social communication skills, reduce anxiety, and enhance cognitive outcomes in children with ASD \cite{dawson2010early, rogers2008evidence}.

This highlights the pressing need for precise and reliable diagnostic tools that are objective, sustainable, and accessible. One promising research direction is the analysis of biofluids such as blood, urine, and saliva. Among these, salivary biomarkers present a particularly attractive, non-invasive alternative, offering a convenient and easily collectable sample type suitable for early detection and monitoring of systemic and neurodevelopmental conditions \cite{gallup2022biomarkers}. Saliva contains a diverse range of biological materials, including over 3,000 proteins, thousands of mRNA transcripts, numerous microRNAs, metabolites, lipids, and other molecules. Importantly, saliva collection is generally less stressful for children with ASD compared to blood draws, enhancing compliance and feasibility in pediatric populations \cite{janvsakova2021potential}. However, due to the heterogeneous nature of ASD and the complexity of biological data, advanced analytical approaches are necessary to interpret these high-dimensional datasets effectively.

In this context, network-based models offer a powerful framework to uncover hidden patterns and relationships within salivary biomarker data. By capturing both structural and semantic connections among biological variables, such approaches can enhance classification performance and contribute to the development of precise, non-invasive, and scalable diagnostic solutions for ASD.

Conventional methods such as decision trees, artificial neural networks, Naive Bayes, \emph{k}-Nearest Neighbor, and Support Vector Machines handle many classification tasks effectively. However, when dealing with significant class overlap or arbitrary data distributions, network-based approaches may offer distinct advantages \cite{silva2012network,carneiro2017organizational}. Traditional methods often rely on physical attributes (e.g., distances, distributions), which may limit their ability to capture deeper semantic relationships within the data.

Deep learning techniques, including convolutional neural networks, excel in identifying high-level patterns across large parameter spaces. However, their performance depends heavily on architectural choices and extensive hyperparameter tuning, leading to high computational costs and a lack of interpretability regarding how decisions are made.

Another limitation of many low-level methods is the assumption that all objects have equal relevance, which may overlook the individual importance of each data point, hindering a deeper understanding of the underlying problem \cite{carneiro2017organizational}.

\begin{figure}[htbp]
    \centering
    \subfloat[\label{fig:simpleexamplea}]{
        \includegraphics[width=0.45\textwidth]{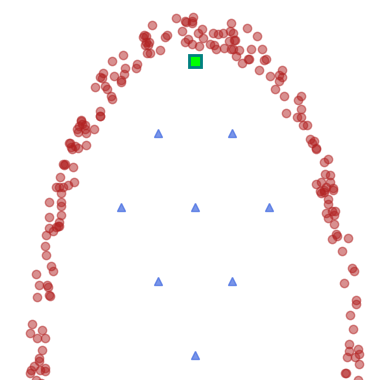}
    }
    \hfill
    \subfloat[\label{fig:simpleexampleb}]{
        \includegraphics[width=0.45\textwidth]{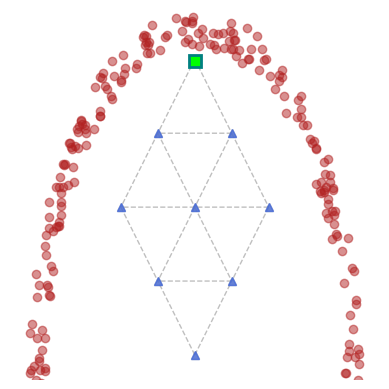}
    }
    \caption[A data set consisting of two classes that form distinct patterns.]
    {Green test object ($\Box$) to be classified between red ($\circ$) or blue ($\triangle$) class.
    (\textbf{a}) Traditional classifiers have difficulties due to the proximity the test object has to the circle class.
    (\textbf{b}) Example of high-level classification: the test object belongs to the diamond pattern.}
    \label{fig:simpleexample}
\end{figure}

In contrast, structural and topological analyses using networks provide an intuitive and semantically meaningful framework. Representing data as a network allows the application of a wide range of concepts and heuristics, capturing both physical and semantic structures \cite{carneiro2017organizational,carneiro2019particle}. For instance, Fig.\ref{fig:simpleexample}(a) illustrates a toy classification scenario where low-level techniques struggle due to their focus on physical attributes while ignoring structural relationships, limiting their ability to correctly classify the green square ($\Box$) as part of the blue triangle ($\triangle$) class.

Network-based methods allow the exploration of topological information in the data configuration, enabling what is known as \emph{high-level classification} \cite{silva2012network}, where semantic relationships are detected beyond simple physical features.

For high-level classification on non-graph data (e.g., attribute vectors, images, text), an essential initial step is the construction of the network. This step is critical, as the graph structure enables the extraction of latent information that enhances classification performance. The effectiveness of high-level classification depends on a representative graph capable of capturing inherent data relationships, as class patterns are directly derived from the network.

Various methods have been proposed for generating graphs from vector-based data in supervised learning contexts, including $k$-associated graphs \cite{bertini2011nonparametric}, $k$-nearest neighbor graphs (kNNG) \cite{carneiro2021complex}, weighted matrix graphs \cite{cupertino2018scheme}, and particle swarm optimized networks (PSONet) \cite{carneiro2019particle}. While these approaches have demonstrated effectiveness in several scenarios, they often rely on heuristic rules or fixed thresholds, which may not capture the most informative connections among data points for complex classification tasks.

In this study, we propose a GA based approach for the structural optimization of networks, aiming to determine the most suitable configuration to represent connections within the network. The motivation for adopting GAs lies in their efficiency in handling discrete optimization problems while naturally reflecting the relational structure of networked data. By representing network construction as a discrete optimization task, GAs can directly manipulate network configurations using binary representations, offering flexibility and adaptability not commonly available in methods that operate on continuous variables.

We hypothesize that GAs can generate networks that are more suitable for classification through importance-based characterization than traditional graph construction methods, such as kNNG, while maintaining competitiveness with state-of-the-art approaches (PSONet).

\subsection{Hypotheses}

Bio-inspired optimization techniques, when combined with network-based measures, can improve high-level classification task by optimizing the underlying graph structure, as demonstrated by approaches like PSONet \cite{carneiro2019particle}. Based on this, we formulate the following hypotheses:

\begin{itemize}
\item \emph{$H_1$: A GA-based graph construction approach provides optimized networks that are competitive with state-of-the-art graph construction method and superior to traditional graph construction methods in classification tasks.}
\item \emph{$H_2$: The proposed GA-based method, GANet, enhances the detection of autism using ATR-FTIR spectra obtained from saliva samples, providing a non-invasive and effective diagnostic approach.}
\end{itemize}

To test $H_2$, we apply our proposed method to the detection of ASD using salivary samples analyzed with attenuated total reflectance Fourier-transform infrared (ATR-FTIR) spectroscopy. In this approach, each spectral sample is represented as a vertex within a network, with edges defined using optimized similarity criteria determined by the GA. Different pre-processing methods, similarity measures, and network metrics are evaluated to assess classification performance on ASD data.

\subsection{Objectives}

The primary objective of this research is to develop and evaluate a GA-based bio-inspired optimization technique for network construction and optimization that enhances high-level data classification through importance-based characterization using complex network measures. This approach directly addresses challenges associated with identifying optimal network structures for classification, offering a novel contribution at the intersection of bio-inspired optimization and complex network analysis, particularly by demonstrating the practical utility of the proposed method in the detection of ASD using non-invasive salivary biomarkers.

The ability to capture and emphasize the relative importance of nodes within the network aligns with the need for efficient models in health diagnostics, addressing the challenges posed by the complexity of high-dimensional spectral data obtained from biological samples.
Specifically, the objectives of this study are:

\begin{itemize}
    \item \emph{Develop a GA-based optimization method:} Design and implement a GA capable of optimizing the topology and connectivity of networks for high-level classification, emphasizing the characterization of node importance to improve the performance of network-based classifiers.
    \item \emph{Validate using ASD salivary spectra data and compare with established methods:} Evaluate the proposed GANet framework against traditional and state-of-the-art classification approaches using saliva samples processed with ATR-FTIR spectroscopy, demonstrating its potential to enhance ASD detection through a sustainable, non-invasive diagnostic pipeline that leverages the strengths of complex network analysis.
\end{itemize}

The remainder of this paper is organized as follows. Section \ref{sec:background} provides a brief overview of the key concepts fundamental to this study. The proposed technique and the main contributions of this work are detailed in Section \ref{sec:ganet_asd}. Section \ref{sec:results} presents the application and results of the proposed method for ASD detection. Finally, Section \ref{sec:conclusion} concludes the paper with final remarks.

\section{Background}\label{sec:background}

Here we present a brief description of the techniques related to this work. Section \ref{sec:ftir} introduces ATR-FTIR as a tool for extracting spectral information from samples. Section \ref{sec:netclass} presents the concept of network-based data classification. Section \ref{sec:importance} discusses importance-based classification methods. Section \ref{sec:gas} introduces genetic algorithms, including a brief explanation of their genetic operators. Finally, Section \ref{sec:ganet} describes the GANet optimization approach, which integrates genetic algorithms and complex networks for enhanced classification performance.

\subsection{Attenuated total reflection-Fourier transform infrared spectroscopy} \label{sec:ftir}

ATR-FTIR spectroscopy is a non-destructive analytical technique widely employed for the characterization of biological samples, including biofluids such as saliva, serum, and urine \cite{movasaghi2008fourier, kazarian2006applications}. This technique enables the acquisition of infrared spectra directly from samples with minimal preparation, providing a rapid and reliable means of obtaining molecular fingerprints based on the vibrational modes of chemical bonds within the sample \cite{kuan2015overview}.

In ATR-FTIR spectroscopy, an infrared beam is directed onto an internal reflection element (IRE) with a high refractive index, typically made of diamond or zinc selenide. When the infrared light undergoes total internal reflection within the IRE, an evanescent wave penetrates a few micrometers into the sample in contact with the crystal surface. This interaction allows for the absorption of specific wavelengths corresponding to the vibrational energy levels of the sample's molecular components \cite{stefan2022ftir}. The resulting spectra provide detailed information about the biochemical composition of the sample, capturing features associated with proteins, lipids, nucleic acids, and carbohydrates \cite{butler2016using}.

The use of ATR-FTIR spectroscopy for the analysis of saliva is particularly advantageous due to its non-invasive nature, minimal sample volume requirements, and the ability to capture comprehensive biochemical profiles without the need for labeling or reagents \cite{poulter2020ftir}. These characteristics make ATR-FTIR a suitable tool for biomedical applications, including disease diagnostics and monitoring, where it has been applied for cancer detection, metabolic disorder assessment, and the evaluation of neurodevelopmental conditions such as ASD \cite{lee2020diagnosing, kazarian2017infrared}.

Moreover, ATR-FTIR spectroscopy generates complex, high-dimensional spectral data that can pose challenges for interpretation using conventional techniques. To effectively extract meaningful patterns from these intricate datasets, advanced analytical methods, including multivariate analysis, machine learning, and network-based classification are essential for revealing subtle biological variations within complex samples. \cite{ellis2012principal, frogley2003fourier}.

\subsection{Network-based data classification} \label{sec:netclass}

Data classification is a central task in machine learning, closely linked to human cognitive processes \cite{carbonell1983overview}. Humans naturally classify objects, whether categorizing people (e.g., colleagues, relatives, classmates) or phenomena such as weather conditions (e.g., sunny, cloudy, rainy). This cognitive ability is influenced not only by physical similarities, such as visual or numerical patterns, but also by the semantic context and relational structures we assign to the elements being classified.

Traditional classification methods typically rely on low-level features, such as distances, densities, or hyperplane separability to assign class labels to data points. While effective in many domains, these techniques may struggle when class boundaries are nonlinear, overlapping, or dependent on relational structures not directly encoded in the feature space.

An alternative approach is to model data as a network, or graph , enabling the use of \emph{topological and structural information} in the classification process. A graph is a mathematical structure defined as $G = (V, E)$, where $V$ is a set of vertices (or nodes), each representing an individual data instance, and $E$ is a set of edges (or links), denoting relationships or similarities between pairs of nodes. In this context, the data instances are embedded in a network based on a similarity function (e.g., Euclidean distance, correlation, or kernel-based metrics), and classification can be performed by analyzing the resulting network structure.

Edges in $E$ can be either weighted or unweighted, and directed or undirected, depending on the nature of the relationship being modeled. For example, in a $k$-nearest neighbor graph (kNNG), each node is connected to its $k$ most similar neighbors, resulting in a sparse, local topology. Alternatively, similarity thresholds or optimization algorithms may be used to define the edge set, allowing for more global or adaptive connectivity schemes.

In network-based classification, also referred to as \emph{high-level classification}, decisions are informed not only by the features of an individual instance but also by its position and role within the network. Nodes can be characterized using centrality measures (e.g., degree, betweenness, closeness), which can reflect the influence or importance of a node in the classification process.

This paradigm is particularly powerful for datasets where complex, non-linear, or hidden relationships exist between data points, such as in biological systems, social networks, and spectroscopic data from biofluids. By leveraging both the local properties of nodes and the global structure of the network, this approach enables richer data representation and more accurate classification outcomes, particularly when integrated with bio-inspired optimization techniques.

\subsection{Importance-based classification} \label{sec:importance}

Classification via importance characterization is a high-level classification technique that individually assesses the importance of data items to determine the label of a new instance \cite{carneiro2018organizational}. Additionally, it leverages both spatial and structural properties by representing data in the form of a graph. In this technique, the concept of \emph{importance} is derived from a centrality measure known as \emph{PageRank}, which characterizes the importance of a given object within a network based on the number of edges directed toward it, such that the more connections a vertex receives, the more important it is considered to be \cite{page1999pagerank}.

In classification via importance characterization, as originally defined in \cite{carneiro2018organizational}, the \emph{importance} of a test item \( y \), denoted by \( \mathcal{I} \), with respect to a class \( l \in \mathcal{L} \), is given by:

\begin{equation}
\mathcal{I}^{(l)}_y = \sum_{j \in \Lambda^{(l)}_y}{\mathcal{I}_j},
\end{equation}

where \( j \in \mathcal{X}_{train} \) denotes a labeled vertex, \( \Lambda^{(l)}_y \) is the set of nodes belonging to class \( l \) to which the test item \( y \) is temporarily connected, and \( \mathcal{I}_j \) represents the \emph{importance} of vertex \( j \).

Classification via importance characterization consists of two main phases. In the training phase, the graph is constructed from the input data (in the form of attribute vectors), after which measures of information flow efficiency and PageRank-based importance are calculated for each vertex. In the testing phase, the process involves the virtual insertion of a test object (query) into the graph by connecting it in a manner that improves information flow efficiency while assigning an importance score to the test object. The object is then assigned to the class corresponding to the component for which it receives the highest importance value \cite{carneiro2018organizational}.

\subsection{Genetic algorithms} \label{sec:gas}

Genetic Algorithms (GAs) are a bio-inspired computational framework for search and optimization that explores the solution space through the abstraction of evolutionary processes and genetic operations, such as reproduction, crossover, and mutation \cite{katoch2021review}. In this framework, a population of candidate solutions is evolved over a predefined number of generations, progressively refining the solutions toward optimal or near optimal configurations.

\begin{figure}[htbp]
\centering
 \includegraphics[width=\textwidth]{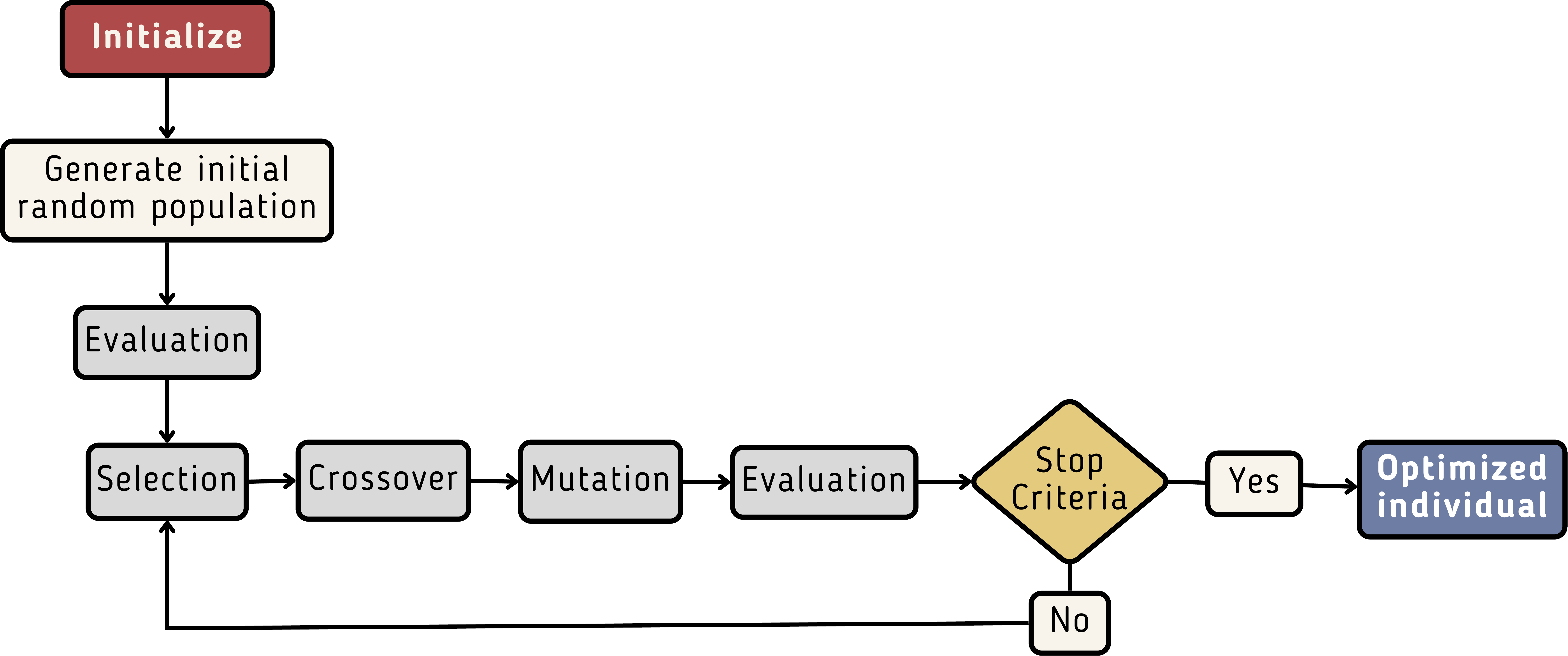}
\caption{Flowchart depicting the main stages of a conventional GA process, from initialization and generation of a random population to iterative steps of evaluation, selection, crossover, and mutation, until meeting stop criteria, resulting in an optimized individual.}
\label{fig:ag}
\end{figure}

Each individual within the population represents a potential solution in the search space, evaluated according to a performance metric known as the fitness function, which quantifies how well a given individual solves the given problem. As illustrated in Fig.~\ref{fig:ag}, the GA process begins with the random initialization of a population of candidate solutions, followed by their evaluation using the fitness function. The optimization proceeds iteratively through the following key steps:

\begin{itemize}
    \item \emph{Selection:} Individuals with higher fitness scores are preferentially chosen to contribute to the next generation, ensuring that promising solutions are retained.
    \item \emph{Crossover:} Selected individuals are combined to generate new offspring, allowing the algorithm to explore new regions of the search space by recombining features from high-performing solutions.
    \item \emph{Mutation:} With a defined probability, random modifications are introduced into the offspring, promoting diversity and enabling the exploration of a broader solution space to avoid premature convergence.
    \item \emph{Reinsertion:} The newly generated individuals are evaluated, and a selection mechanism determines which individuals will form the next generation, balancing exploitation of high-quality solutions with exploration of new candidates.
\end{itemize}

This evolutionary cycle is repeated until a stopping criterion is met, such as reaching a maximum number of generations or achieving a satisfactory fitness level. Ultimately, the individual with the highest fitness score is returned as the best solution. The adaptability and global search capabilities of GAs make them a suitable tool for tackling complex optimization problems, particularly in scenarios involving discrete search spaces and high-dimensional parameter configurations.

\begin{figure}[htbp]
\centering
 \includegraphics[width=\textwidth]{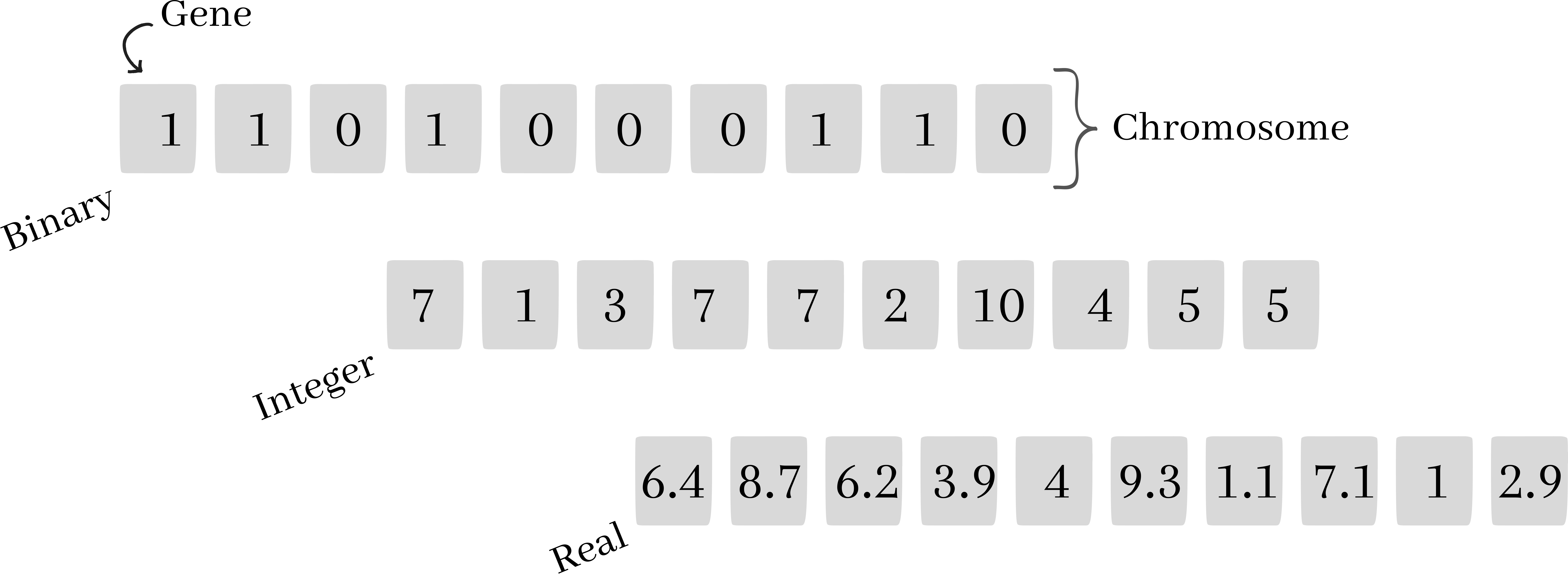}
\caption{The three principal individual representations: binary, integer and real.}
\label{fig:representation}
\end{figure}

One of the great challenges in the design of a GA refers to the representation and evaluation of individuals, as well as the consequent choice of methods adopted in the genetic operators of selection, crossover, mutation and reinsertion \cite{katoch2021review}. The individual representations are mainly through the binary, integer, and real numbers. In the Fig. \ref{fig:representation} examples of them can be seen.

\subsubsection{Genetic operators}

\begin{figure}[htbp]
\centering
 \includegraphics[width=.7\textwidth]{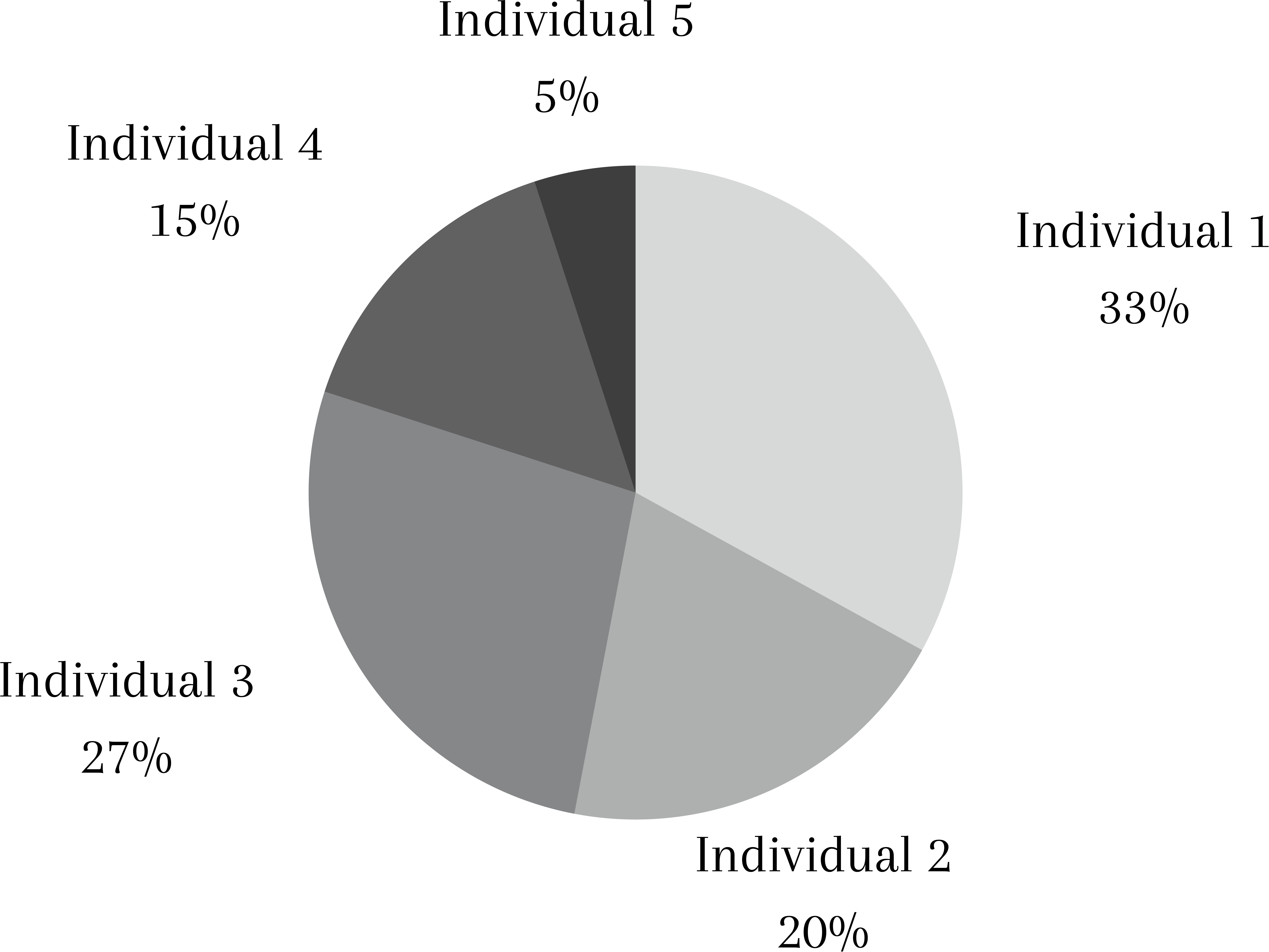}
\caption[Roulette selection technique]{Roulette selection technique: selection strategy for distinguishing the merits of the data based on their fitness. Because \emph{Individual 1} has the highest fitness, it receives the largest piece of the roulette wheel.}
\label{fig:simple-roulette}
\end{figure}

The \emph{selection} of the best individuals for recombination is fundamentally stochastic and can be done through methods such as Roulette, Ranking, Tournament, Truncation and among others \cite{yadav2017comparative}. In the \emph{roulette} selection method, illustrated in Fig.\ref{fig:simple-roulette}, each individual has the probability of being selected according to its associated fitness.

\begin{figure}[htbp]
    \centering
    \subfloat[\textit{Two points crossover}\label{fig:crossover2points}]{
        \includegraphics[width=0.6\textwidth]{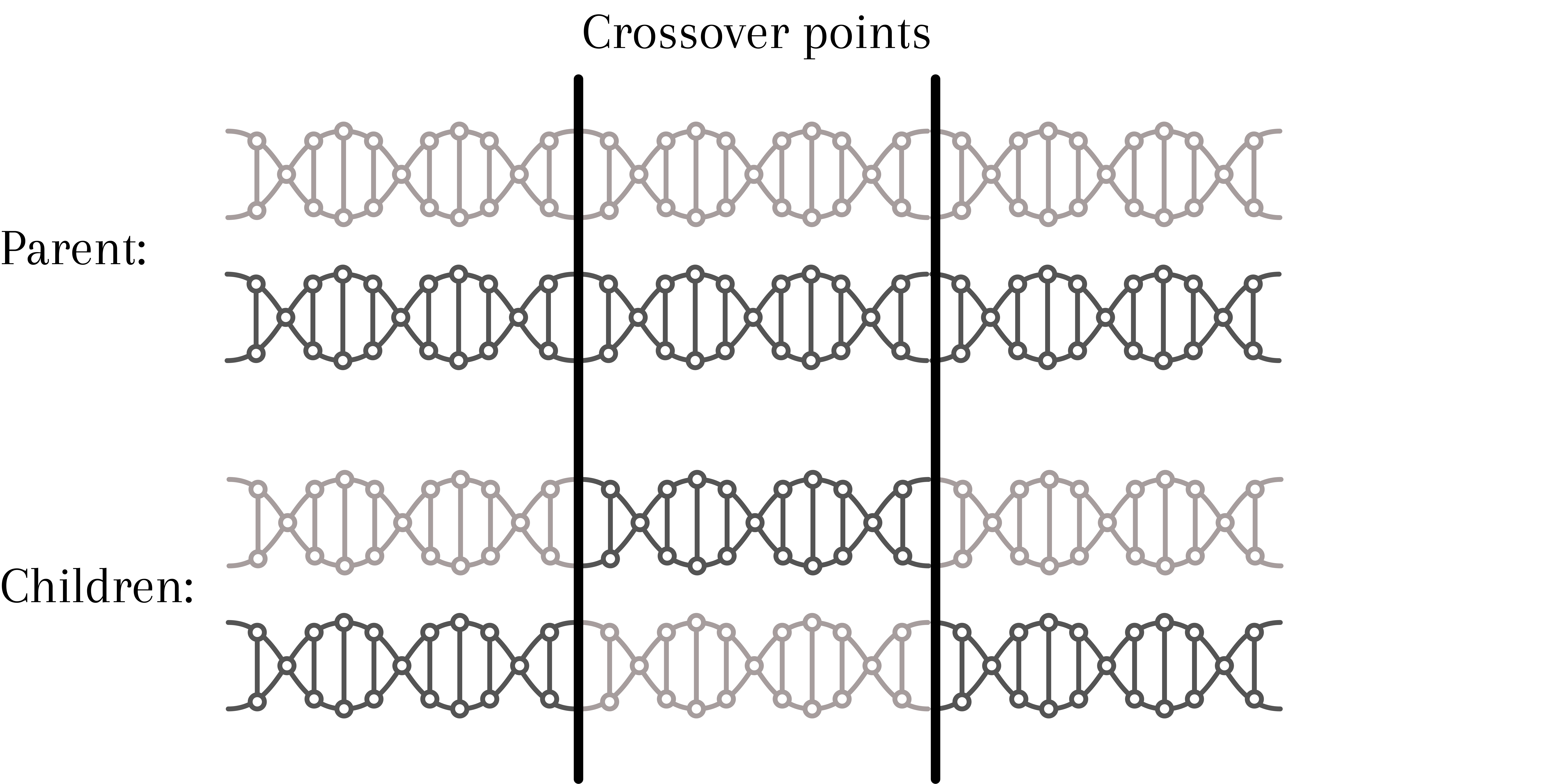}
    }
    \hfill
    \subfloat[\textit{Uniform crossover}\label{fig:uniform}]{
        \includegraphics[width=0.35\textwidth]{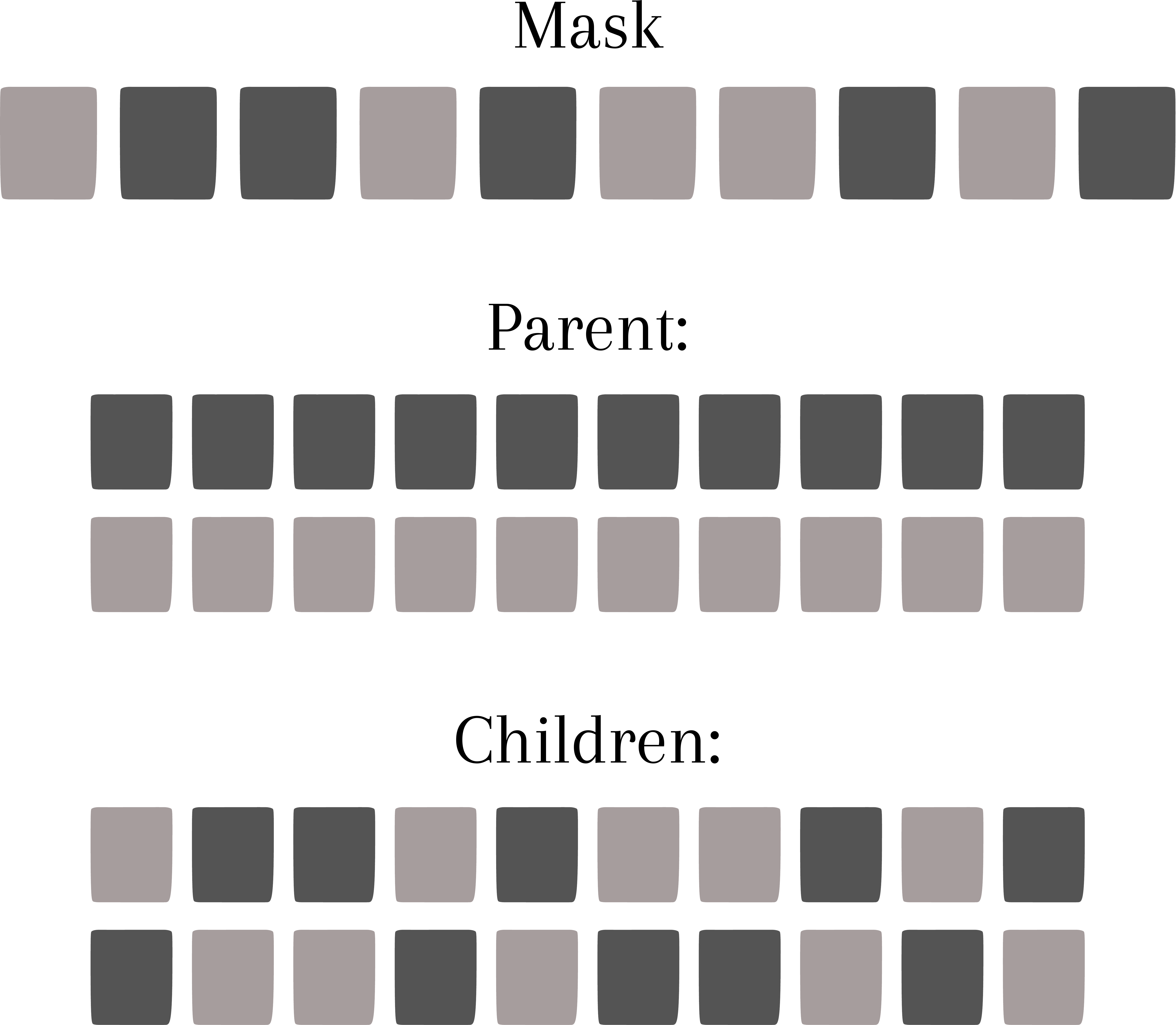}
    }
    \caption[Crossover techniques: two points crossover and uniform crossover]
    {Crossover techniques: \textbf{(a)} \textit{two points crossover}: two crossover points are randomly selected, and the individuals swap the bits between these points; \textbf{(b)} \textit{uniform crossover}: a mask indicates that the bit is copied from the first parent, otherwise from the second parent.}
    \label{fig:simple-crossover-2}
\end{figure}

In \emph{crossover} the selected individuals are used for genetic recombination and thus will generate the children that will compose the total population $(Tp+Cr)$. Each child is made up of their parents' material and the crossover technique determines which parent’s genes will be used. Cyclic Crossover, PMX, Two Point Crossover and Uniform Crossover are examples of crossover techniques  \cite{kora2017crossover}. The below Fig.\ref{fig:simple-crossover-2}  illustrates both two points and uniform crossover.

Biologically, \emph{mutation} means an abrupt change in the characteristics of a gene on a chromosome. In this algorithm phase, the children generated can undergo genetic variations according to a predefined probability. These genetic variations can occur, depending on the representation of the individual, by the alteration of a randomly chosen gene or by the random replacement of one gene by another. The mutation aims to add diversity and increase exploration of the search space \cite{mirjalili2019genetic}.

In \emph{reinsertion}, the selection takes place among the total individuals $(Tp+Cr)$ which will compose the next generation $(Tp)$ based on the fitness of the individuals. Examples are Ordered Reinsertion and Pure Reinsertion \cite{mirjalili2019genetic}.

\subsection{GANet optimization} \label{sec:ganet}

The GANet method is divided into two main phases: training and testing. During the training or optimization phase, the GA is employed using the training data to obtain an optimized network configuration. In the testing phase, the best solution obtained during training is used to construct the network, which now incorporates the test data, leveraging the classification via importance characterization technique and the optimized network for the classification process.

Formally, the GA manipulates a population of individuals \( P = \{I_1, I_2, \ldots, I_m\} \), where each individual \( I_i \in P \) is defined as:
\begin{equation}
    I_i = \{v_1, v_2, \ldots, v_n\},
\end{equation}
where \( v_i \in I_i \) represents the connections of a given vertex \( v_i \) (associated with an object \( x_i \in X \)), defined as:
\begin{equation}
    v_i = \{e_{i1}, e_{i2}, \ldots, e_{iq}\},
\end{equation}
where \( j \in \{1,2,\ldots,q\} \) denotes the \( q \) possible connections of \( v_i \), and \( e_{ij} \in \{0,1\} \) indicates the presence or absence of a connection from vertex \( v_i \) to its neighboring vertex \( Map_{ij} \) in the network, as illustrated in Fig. \ref{fig:map-all}. The neighbors of each vertex \( v_i \) are defined based on the \emph{MapAll} mapping heuristic, originally proposed in \cite{carneiro2019particle}, which defines the \( Map \) matrix following these steps:
\begin{enumerate}[label=\alph*)]
    \item Compute the similarity between each pair of data items;
    \item For each vertex \( v_i \), select its \( q \) most similar vertices;
    \item Given \( 1 \leq z \leq q \), create the \( Map_{n \times q} \) matrix such that:
\end{enumerate}
\begin{equation}
Map_{iz} =
    \begin{cases}
        v_z & \quad \text{if } l_i = l_z\\
        \emptyset & \quad \text{otherwise.}
    \end{cases}
\end{equation}
Note that \( Map_{iz} \) is empty if vertex \( v_i \) does not belong to the same class as \( v_z \). It is also important to highlight that, unlike the continuous optimization method presented in \cite{carneiro2019particle}, the GA developed here performs optimization in a discrete solution space.

Initially, the data are split into training, validation, and testing sets, and a random population of \( m \) individuals is generated using the MapAll mapping heuristic based on the training data. Each individual in the population undergoes the evolution process, including selection, crossover, and mutation over \( m \) generations. During the optimization phase, each individual \( \mathcal{I}_k \) is evaluated using a fitness function on the validation set. Each \( \mathcal{I}_k \) is converted into a network \( \mathcal{G}_k = \{\mathcal{V}_k, \mathcal{E}_k\} \), where \( \mathcal{V}_k = \{1,\ldots,n\} \) represents the vertices associated with each data item, and \( \mathcal{E}_k \) represents the edges between these vertices. The adopted fitness function is the \emph{classification via importance characterization} proposed in \cite{carneiro2018organizational}.

In the testing phase, the best individual obtained at the end of the \( m \) generations is used to construct the network for high-level classification using the test dataset.

\section{GANet high-level classification for ASD detection} \label{sec:ganet_asd}

In high-level classification, the primary objective is to examine the properties and metrics of complex networks to characterize the emergence of patterns in the classification of individual test items. This methodology provides several advantages over conventional low-level techniques, including the elimination of parameter dependency, the ability to detect overlapping classes, and the capacity to capture intricate spatial, topological, and functional relationships within the dataset.

The kNNG method is one of the most commonly used graph construction techniques. This approach generates a directed graph in which each vertex is connected to its \emph{k} nearest neighbors, provided that the corresponding objects belong to the same class. Another widely employed method is the $\epsilon$-radius neighborhood (radius-$\epsilon$), which constructs an undirected graph where connections between vertices of the same class are established based on a distance threshold $\epsilon$ \cite{fernandes2023network}. These methods construct graphs based on input data represented as attribute vectors, inherently making strong assumptions about the data to varying extents. Specifically, they assume that relationships between the data points can be mapped through a uniform number of connections (\emph{k} or $\epsilon$) between vertices or network components \cite{carneiro2019particle}.

An alternative approach to overcoming these limitations is through the structural optimization of networks. In this work, we propose GANet, a framework that leverages a GA to manipulate network configurations through a binary representation. This encoding allows the optimization process to explore a significantly smaller, finite configuration space compared to the literature benchmark, PSONet \cite{carneiro2019particle}.

The proposed method is structured into two primary phases: training and test. In the training phase, also referred to as the optimization phase, the GA is utilized to derive an optimal configuration of the network based on the training data. Subsequently, during the test phase, the optimal solution identified in the training phase is employed to construct the network that processes the test data. The classification is then performed using the technique of importance characterization in conjunction with the optimized network, enabling the effective categorization of the data.

Importance characterization evaluates the significance of each vertex based on a specific network measure. In this work, we employed the degree  measure \cite{fernandes2023data} to assess vertex importance.  Degree centrality, a straightforward measure that assesses a vertex's importance based on its direct connections, offers notable advantages in computational efficiency and robustness, thus enhancing its suitability for large-scale datasets and real-time applications where computational resources may be constrained. Additionally, degree centrality has the potential to deliver comparable predictive performance while maintaining lower computational complexity, making it a valuable alternative for high-level classification tasks \cite{fernandes2023data}.

\begin{figure}[htbp]
    \centering
    \includegraphics[width=\textwidth]{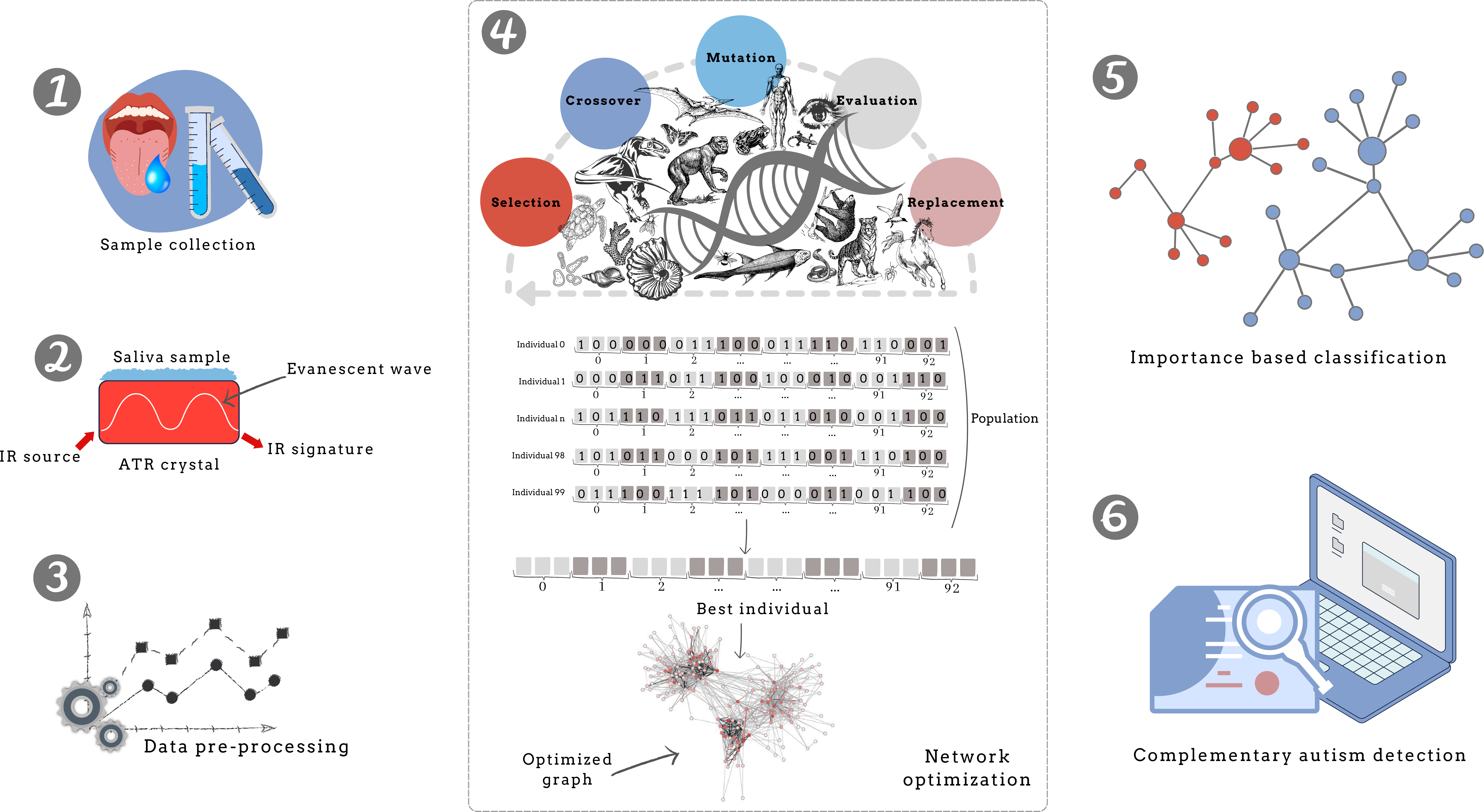}
    \caption[GANet overall concept]{GANet overall concept. (1) Saliva collection is obtained. (2) The samples are processed through the spectrometer. (3) Spectra data is pre-processed. (4) GANet evolves better-suited individuals. (5) High level data classification occurs. And, (6) GANet complements autism detection. In step (4) there is an example of the proposed encoding strategy which represent the network. The final individual represents the evolved network, which is considered to be the optimal candidate solution. The network's encoding comprises the interconnections among each salivary spectra. For each vertex, its codes are the map of the \emph{q} value of possible neighbors. The neighbors' potential connections are represented by the numbers 3 and 5.}
    \label{fig:proposed-ga-net-ftir}
\end{figure}

The Fig. \ref{fig:proposed-ga-net-ftir} illustrates the fundamental concept underlying the proposed method.

\begin{enumerate}
\item \emph{Sample collection}: Saliva samples are collected from individuals. This method is non-invasive and cheap, which is advantageous, especially for pediatric diagnostics, as it minimizes discomfort compared to blood collection.

\item \emph{FTIR spectroscopy}: The saliva samples undergo processing through FTIR spectroscopy, a process used to capture the unique infrared (IR) signatures of each sample that represent their biochemical composition. 

\item \emph{Data pre-processing}: In the data pre-processing stage, steps are taken to eliminate noise and irrelevant features, optimizing the dataset for subsequent analysis. This study specifically examines three pre-processing techniques: Smoothing using the Savitzky-Golay filter, Differentiation, and Spectrum Truncation, each evaluated for their effectiveness in enhancing data quality.

\item \emph{Network optimization}: Following data pre-processing, GANet evolves better-suited individuals for the classification via characterization of importance. First, a population of candidate solutions is initialized randomly and then evaluated using a fitness function. The optimization process then begins, running for a specified number of generations or until another stopping condition is met. This process includes the following stages:

\begin{enumerate}
\item \emph{Selection}: The best individuals or solutions, higher-evaluated, are given preference for participating in the crossover phase.
\item \emph{Crossover}: Selected individuals contribute combining genetic information to produce offspring that may inherit beneficial features from each parent.
\item \emph{Mutation}: Minor alterations with some probability are introduced to the newly generated individuals to promote diversity and explore new potential solutions, allowing for global exploration of the search space.
\item \emph{Evaluation}: The resulting individuals are then evaluated using the fitness function.
\item \emph{Reinsertion}: Poor-performing individuals are replaced by new ones to refine the population.
\end{enumerate}

At the end of this process, the best-evaluated and evolved individual is returned. This aids in the initial classification by focusing on the most relevant data points that may improve classification task.

\item \emph{Importance-based classification}: The classification process is further optimized by applying the resultant network to categorize new data through importance characterization. This approach involves the virtual integration of patient data within the model's components based on improving the efficiency of the information flow. Subsequently, an importance value is assigned to each data input, enabling classification into the component class where the data exhibits the highest importance score. This method ensures that classification aligns with the component best suited to capture the critical attributes of the patient data.

\item \emph{Complementary ASD detection}: The GANet serves as a complementary ASD detection tool, supporting doctors in their diagnostic efforts. The process aims to accurately identify indicators of autism, enhancing clinical assessments with a non-invasive and data-driven approach. By providing insights into subtle patterns that may be challenging to detect through traditional methods, GANet aids healthcare professionals in making more informed and timely diagnoses, ultimately improving care for individuals on the autism spectrum.
\end{enumerate}

This systematic method allows for a high-level classification framework that leverages complex network analysis and bio-inspired optimization techniques to improve diagnostic accuracy and efficiency.

\section{Experimental results} \label{sec:results}

This section provides the results of experiments designed to validate and assess the performance of the proposed GANet framework for the detection of ASD using salivary ATR-FTIR spectral data.

\subsection{Dataset description}
\label{sec:spectra-processing}
The database was acquired with the authorization of the Research Ethics Committee of the Federal University of Uberlândia, in accordance with protocol 249.200.9.

This study had a total of 159 samples from 53 participants, consisting of 19 verified cases of ASD and 34 controls. The control group was identical to the ASD group in terms of age, gender, and ethnicity, and both groups contained male and female patients. Furthermore, 26.31\% of the participants exhibit attention deficit/hyperactivity disorder in addition to ASD. There are three processed saliva samples per participant, resulting in triplicate data.

An FTIR spectrum provides a detailed and consistent molecular representation of the chemical composition of a sample's unique biological signals. Hence, the combination of FTIR spectral fingerprints and multivariate spectral analysis is frequently employed to characterize and identify saliva biofluid. Usually, the study of infrared spectra involves pre-processing the spectra and designing models, such as clustering or classification.

In the pre-processing stage, it is customary to exclude spectra that exhibit a poor signal-to-noise ratio and high noise. Additionally, spectra displaying low intensity of a pertinent peak, such as the amide I peak, or spectra displaying high intensity of an undesired or irrelevant peak(s), such as water vapor in samples, are also disregarded.

\begin{figure}[hbtp]
    \begin{center}
        \subfloat[\label{fig:raw}]{
            \includegraphics[width=0.32\linewidth,keepaspectratio]{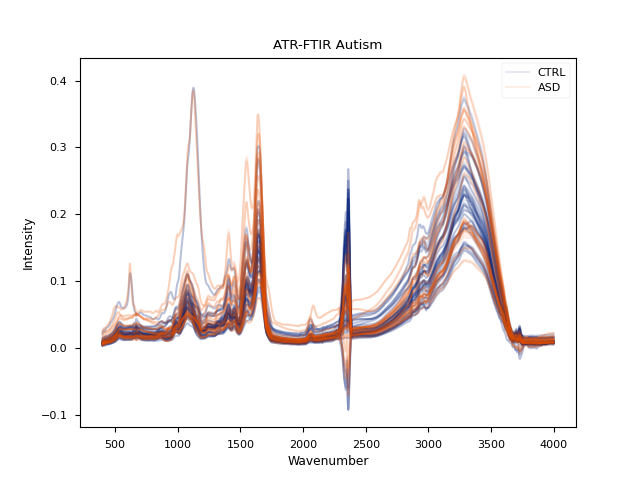}
        }
        \subfloat[\label{fig:sdn}]{
            \includegraphics[width=0.32\linewidth,keepaspectratio]{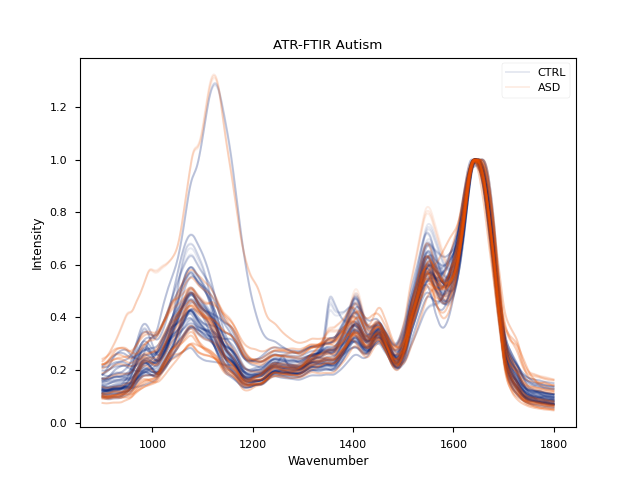}
        }
        \subfloat[\label{fig:sdnt}]{
            \includegraphics[width=0.32\linewidth,keepaspectratio]{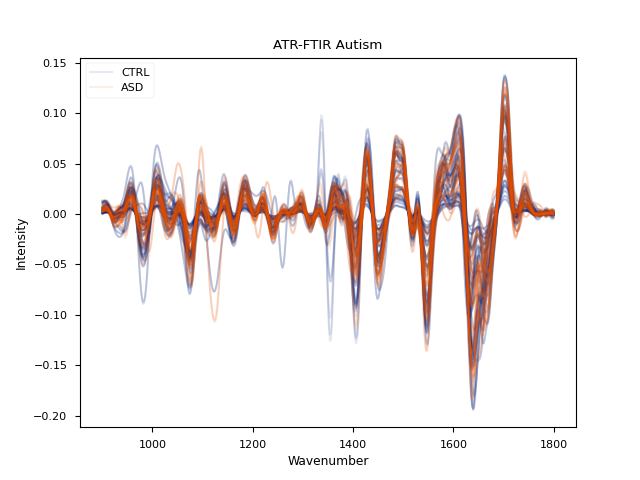}
        }
        \caption[Data pre-processing techniques plot of saliva samples]
        {Data pre-processing techniques plot of saliva samples where blue indicates control data and orange ASD data.
        \textbf{(a)} Raw;
        \textbf{(b)} Normalization and truncation;
        \textbf{(c)} Smoothing, differentiation, normalization, and truncation.}
        \label{fig:pre-processing-ftir}
    \end{center}
\end{figure}

This study involved the pre-processing of FTIR spectra data through the use of area normalization, smoothing, differentiation, and truncation techniques. Fig. \ref{fig:pre-processing-ftir} depicts the alterations in the data subsequent to the implementation of each pre-processing approach. Below are explanations of each of these pre-processing techniques:

\begin{enumerate}
\item \emph{Normalization by the peak of amide (amide I)}: Discrepancies in thinness or concentration might sometimes be the most noticeable cause of spectrum variation across samples, masking the important metabolic changes. In order to mitigate these effects, the spectra are adjusted to meet a certain requirement. The application of normalization to a specific peak is possible. In this study, the process of normalizing the spectra is performed by identifying the peak of amide 1, which is the greatest value within the infrared band range spanning from 1660 cm$^{-1}$ to 1630 cm$^{-1}$. This particular region has promise for protein characterisation.

\item \emph{Smoothing with Savitzky-Golay filter}: The Savitzky-Golay filter is employed in signal processing to diminish noise in a signal and enhance the smoothness of its trend. The filter calculates a polynomial fit for each window based on the polynomial degree and window size. It is important to note that the process of smoothing might potentially eliminate important information in spectra, but it can also introduce distortions into the spectral data.

\item \emph{Differentiation}: Both baseline correction and the separation of overlapping bands were employed. Typically, the gradients of the pertinent spectral bands exhibit much greater magnitudes compared to the preexisting baselines, and the process of differentiation is employed to amplify these disparities.

\item \emph{Spectrum truncation}: In order to mitigate the impact of noise and outliers on the generalization performance of trained models, the spectra are condensed to a range spanning from 900cm$^{-1}$ to 1800cm$^{-1}$.
\end{enumerate}

\subsection{Encoding} 

\begin{figure}[htbp]
    \centering
    \includegraphics[width=\textwidth]{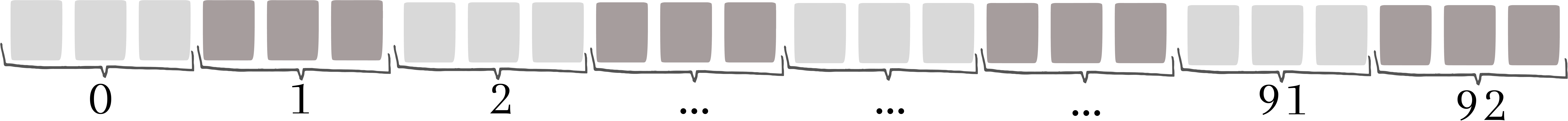}
    \caption[GA-encoded salivary FTIR data for autism detection.]{Individual encoding for autism detection on salivary FTIR data through GAs. In this example, the genome size is 93$\times$3. There are 93 positions for 93 salivary spectra samples. There are three possibilities for establishing connections with surrounding vertices. The \emph{q} parameter pre-determines the number of potential connections.}
    \label{fig:individual-ga-ftir}
\end{figure}

The encoding strategy used in this study represents each ATR-FTIR saliva spectrum as a vertex in the network. After processing saliva samples using ATR-FTIR, the resulting spectra are directly mapped to vertices within the constructed network. Within the GA framework, each individual in the population defines the connections (edges) between these vertices, determining the network structure. Fig.~\ref{fig:individual-ga-ftir} illustrates the detailed encoding scheme used for each individual.

A total of 159 salivary samples from 53 participants, with three processed saliva samples per participant (triplicates), were divided into training (93 samples), validation (33 samples), and test (33 samples) sets, ensuring that all replicates from a single participant were assigned to the same subset. This preserves subject independence and provides a reliable assessment of the model’s real-world generalization. The validation set was reserved to evaluate the model's generalization performance on unseen data, ensuring independence from the training process. For the training set of 93 samples, the corresponding network consists of 93 vertices, and each individual (genome) in the GA population contains 93 chromosomes, one for each vertex. Each genome includes a mapping structure that specifies potential connections to other vertices in the network, with the size of this mapping determined by the predefined parameter \emph{q}, which defines the number of candidate connections considered for each vertex.

\begin{figure}[htbp]
    \centering
    \includegraphics[width=\textwidth]{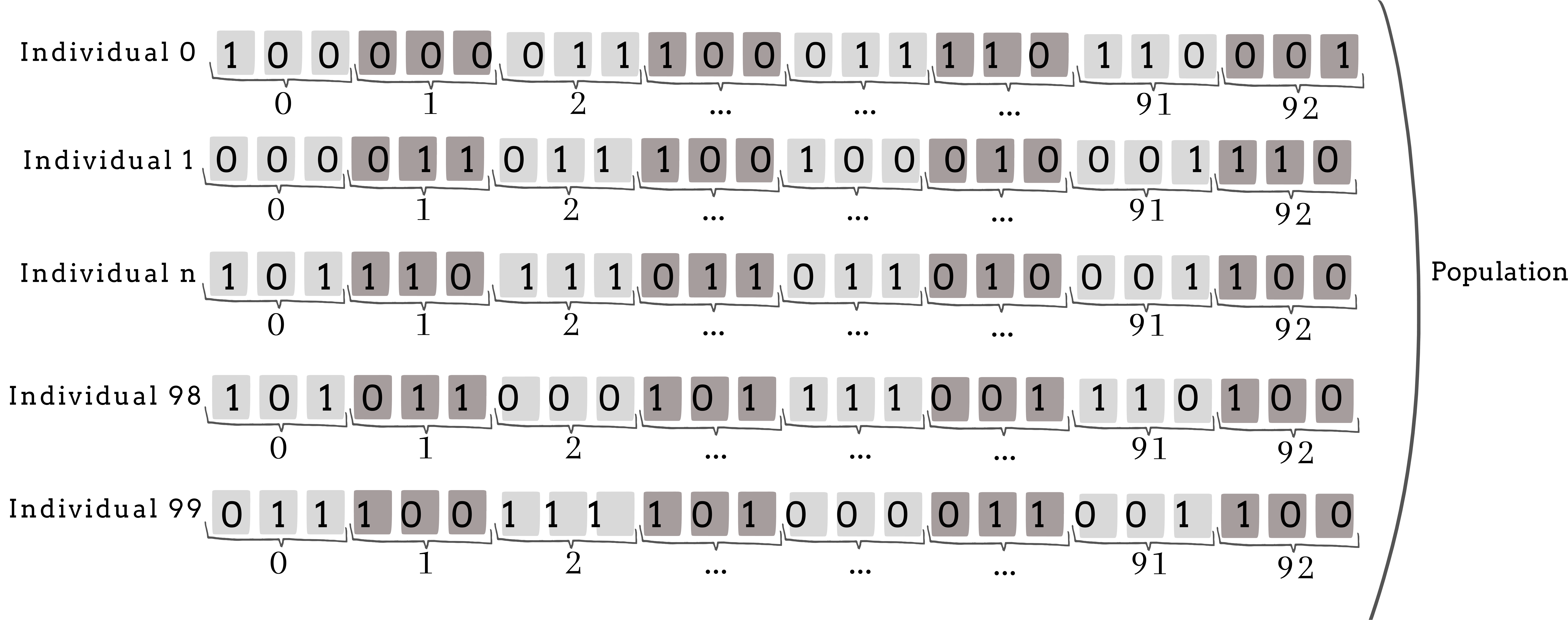}
    \caption[Initial generation of individuals example]{The procedure starts by initializing a population of encodings that is randomly selected, consequently constituting the initial generation of individuals. Example with a population size of 100. Individuals have a genomic size of 93x3.}
    \label{fig:population-init-ftir}
\end{figure}

The study included a population size of 100 individuals, resulting in a total of 100 potential solutions. The population is first generated using random values, where a value of $1$ indicates the existence of an edge connecting the neighboring nodes, while a value of $0$ indicates the lack of such connection. The Fig. \ref{fig:population-init-ftir} depicts an illustrative instance of a population of potential solutions that have been randomly initiated.

\begin{figure}[htbp]
    \centering
    \includegraphics[width=\textwidth]{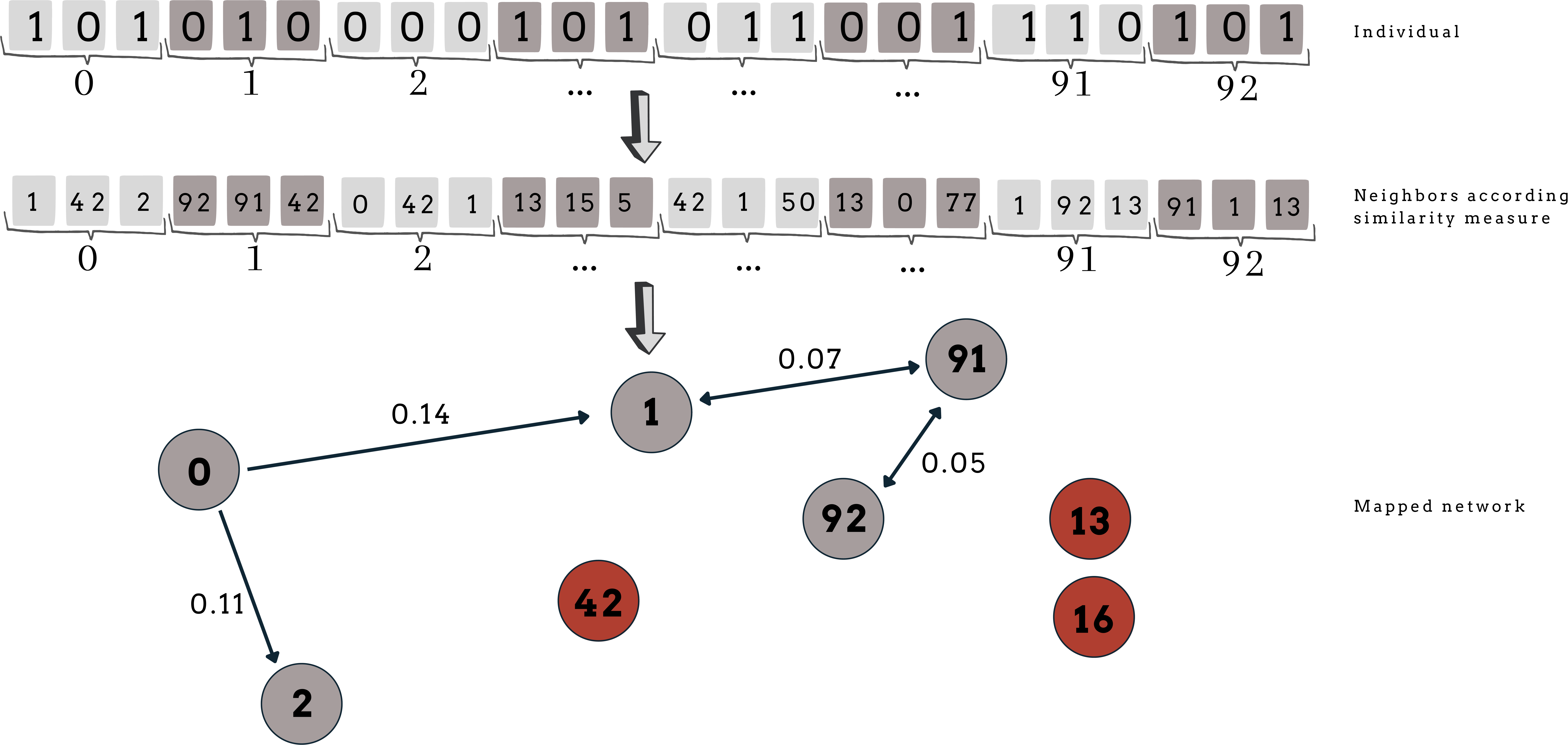}
    \caption[Individual encoding representing a network]{An individual encoding network. The selection of neighbors is determined by a similarity metric. The occurrence of the link is a stochastic process. It is important to acknowledge that in order for connections to exist, the vertices must possess same target classes. An illustrative instance is provided above; vertex 92 is adjacent to vertex 13. Although there is a representation of an association between them in the individual, they do not establish a connection due to their membership in distinct classes.}
    \label{fig:population-map-ftir}
\end{figure}

Fig.~\ref{fig:population-map-ftir} illustrates an example of the mapping process used in this study, with \emph{q} set to 3 in this scenario. A similarity metric, either Euclidean or Cosine distance, is employed to determine the likelihood of vertices being considered neighbors. Each individual in the GA population maintains a dictionary structure that specifies potential connections for constructing the network.

In the illustrated example, vertex 0 forms connections with its two most similar neighbors, vertex 1 connects with one neighbor, and this process continues for each vertex in the dataset. Connections within the network are only established between vertices belonging to the same class, ensuring that the network structure aligns with the class-based organization required for classification tasks. For instance, although vertices 92 and 13 are close in the similarity ranking, they do not form a connection in the network because they belong to different classes.

This restriction preserves the integrity of the classification structure by preventing connections that could introduce inconsistencies within the network, ensuring that the network representation respects class boundaries while leveraging similarity information to define meaningful connections.

\begin{figure}[htbp]
\centering
\includegraphics[width=1\textwidth,keepaspectratio]{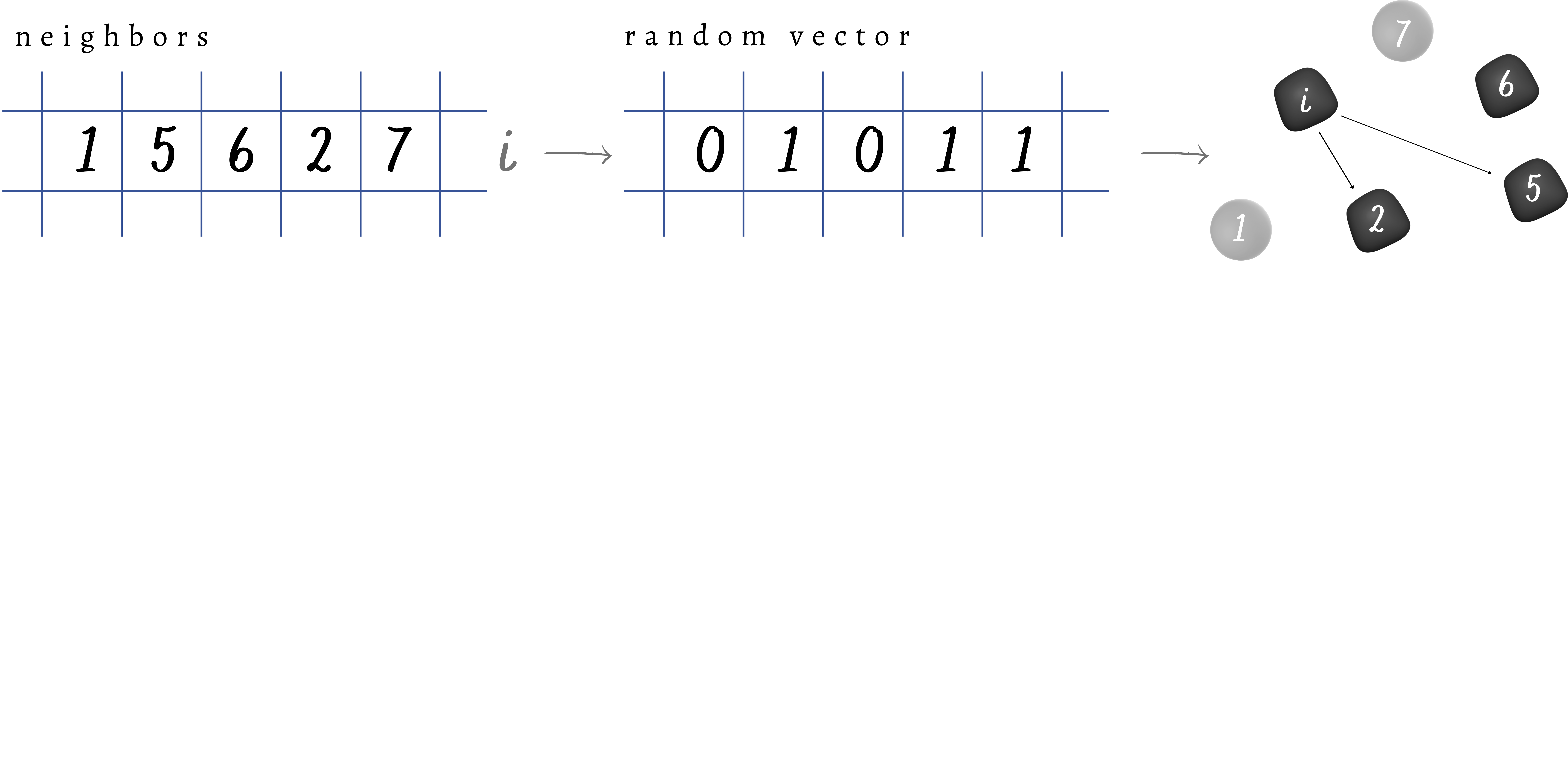}
\caption[Illustrative example of \emph{Map-all}]{Illustrative example of \emph{Map-all} where from a random vector vertices of the same class are connected. In this example there is no connection between \emph{i} and 7 vertices as they are of different classes, however, even if \emph{i} and 6 vertices are of the same class there is no connection between them because there is no such connection in \emph{Map-all} vector.}
\label{fig:map-all}
\end{figure}

In a formal way, the GA manipulates a population of individuals $P = \{I_1, I_2, \ldots, I_m\}$, in which each individual $I_i \in P$ is denoted by:
\begin{equation}
    I_i = \{v_1, v_2, \ldots, v_n\},
\end{equation}
where $v_i \in I_i$ represents the connections of a given vertex $v_i$ (associated with an object $x_i \in X$) defined by:
\begin{equation}
    v_i = \{e_{i1}, e_{i2}, \ldots, e_{iq}\},
\end{equation}
where $j \in \{1,2,\ldots,q\}$ denotes the possible $q$ connections of $v_i$ and $e_{ij} \in \{0,1\}$ the existence or absence of a given connection from vertex $v_i$ to neighboring vertex $Map_{ij}$ in the network, as illustrated in Fig. \ref{fig:map-all}. The neighbors of each vertex $v_i$ are defined based on the mapping heuristic \emph{MapAll}, originally proposed in \cite{carneiro2019particle} and which defines the matrix $Map$ through the following steps:
\begin{enumerate}[label=\alph*)]
    \item Calculate the similarity between each pair of data items;
    \item Select for each vertex $v_i$ its most similar $q$ vertices;
    \item Given that 1 $\leq z \leq q$, create the matrix $Map_{n \times q}$ such that:
\end{enumerate}
\begin{equation}
Map_{iz} =
    \begin{cases}
        v_z & \quad \text{se } l_i = l_z\\
        \emptyset & \quad \text{otherwise.}
    \end{cases}
\end{equation}
where, $ l \in \mathcal{L}$ is the instance label.
Note that $Map_{iz}$ is empty if the vertex $v_i$ does not belong to the same class as $v_z$. From the formulation presented, it is also important to note that, unlike the continuous optimization method presented in \cite{carneiro2019particle}, the GA developed here performs the optimization in a discrete space of solutions.

After randomly initializing the population with the predefined size using the proposed encoding strategy, each individual’s fitness is evaluated during the evolutionary process. The GA framework is designed to optimize the accuracy of the importance-based classifier, which is used as the fitness function to determine the best individuals to advance to the next generation.

Parent selection is based on fitness values, after which genetic operators, crossover and mutation, are applied to generate new offspring. The environmental selection step combines the parent and offspring populations, selecting the individuals that will form the next generation.

This evolutionary process is repeated over a specified number of generations, resulting in the identification of the best-performing individual. In this study, we employed tournament selection for parent selection, two-point crossover, and ordered and pure reinsertion strategies. As demonstrated in \cite{fernandes2023network}, these operators statistically outperformed kNNG and other evaluated configurations.

\subsection{Experimental setup}

In our experiments, we employed the four most optimal GA configurations identified in \cite{fernandes2023network}. These configurations are defined by the parameters $\gamma$, which controls the trade-off between low and high level features, the neighborhood size \emph{q} used in the network construction, and the choice of selection, reinsertion, and crossover operators within the evolutionary process. The specific configurations, GANet-C, GANet-E, GANet-G, and GANet-K, are detailed in Table~\ref{tab:configsstudy}.

All configurations consistently employ tournament selection and two-point crossover for parent selection and offspring generation, respectively. The primary variations concern the reinsertion strategy, either pure or ordered, and the parameters $\gamma$ and \emph{q}.

The configurations were chosen for this study due to their superior performance in optimizing network structures for high-level classification \cite{fernandes2023network}. These setups are now evaluated in the context of Autism Spectrum Disorder detection using ATR-FTIR saliva spectra, enabling a focused investigation into the adaptability and robustness of GA-based network optimization methods.

Additionally, besides the Euclidean similarity metric, we also evaluated the Cosine similarity metric in our experiments. 

\begin{table}[htbp]
\centering
\caption{Different configurations under investigation consist of the parameters $\gamma$, \emph{q}, and the reinsertion operator.}
\begin{tabular}{l c c l}
\toprule
Configuration & $\gamma$ & \emph{q} & Reinsertion  \\
\midrule
GANet-C & 1 & 3 & Pure        \\
GANet-E & 1 & 5 & Ordered    \\
GANet-G & 1 & 5 & Pure        \\
GANet-K & 2 & 3 & Pure        \\
\bottomrule
\end{tabular}
\label{tab:configsstudy}
\end{table}

We evaluated the LDA classifier, a method commonly used for FTIR data analysis \cite{morais2020tutorial}, alongside the SVM classifier, which serves as a benchmark in supervised learning and is recognized as a state-of-the-art low-level classifier in ATR-FTIR spectroscopy studies \cite{deiss2020tuning}.

To broaden our investigation, we further included eight deep learning algorithms in our experiments. The architectural details and optimization hyperparameters for these models are provided in Table~\ref{tab:Hyperparameter-deep-architecture} and Table~\ref{tab:Hyperparameter-deep}, respectively. Due to hyperparameter incompatibilities, TWIESN was not included in these tables. Instead, following the methodology described in \cite{tanisaro2016time}, we implemented TWIESN separately and performed a grid search on an unseen 20\% validation split of the training data to optimize its three key hyperparameters: reservoir size, sparsity, and spectral radius.

\begin{table}[htbp]
\caption[Hyperparameters of the architecture for deep learning approaches.]{Hyperparameters of the architecture for deep learning approaches.}
\label{tab:Hyperparameter-deep-architecture}
\begin{adjustbox}{width=\textwidth}
\begin{tabular}{l l l l l l l l l}
\toprule
{} & {Architecture} &          &           &             &           &           &            &              \\ \cmidrule(l){2-9} 
\multirow{-2}{*}{{Methods}} & {\#Layers}     & {\#Conv} & {\#Invar} & {\#Norm} & {Pooling} & {Feature} & {Activate} & {Regularize} \\ \cmidrule(r){1-9}
{MLP}                       & {4}            & {0}      & {0}       & {None}      & {None}    & {FC}      & {ReLU}     & {Dropout}    \\
{FCN}                       & {5}            & {3}      & {4}       & {Batch}     & {None}    & {GAP}     & {ReLU}     & {None}       \\
{ResNet}                    & {11}           & {9}      & {10}      & {Batch}     & {None}    & {GAP}     & {ReLU}     & {None}       \\
{Encoder}                   & {5}            & {3}      & {4}       & {Instance}  & {Max}     & {Att}     & {PReLU}    & {Dropout}    \\
{MCNN}                      & {4}            & {2}      & {2}       & {None}      & {Max}     & {FC}      & {Sigmoid}  & {None}       \\
{t-LeNet}                   & {4}            & {2}      & {2}       & {None}      & {Max}     & {FC}      & {ReLU}     & {None}       \\
{MCDCNN}                    & {4}            & {2}      & {2}       & {None}      & {Max}     & {FC}      & {ReLU}     & {None}       \\
{Time-CNN}                  & {3}            & {2}      & {2}       & {None}      & {Avg}     & {Conv}    & {Sigmoid}  & {None}       \\
\bottomrule
\end{tabular}
\end{adjustbox}
\end{table}

\begin{table}[htbp]
\caption[Hyperparameter optimization for deep learning approaches.]{Hyperparameter optimization for deep learning approaches.}
\label{tab:Hyperparameter-deep}
\begin{adjustbox}{width=\textwidth}
\begin{tabular}{l l l l l l l l}
\toprule
{} & {Optimization} &             &           &          &         &                 &         \\ \cmidrule(l){2-8} 
\multirow{-2}{*}{Methods} & {Algorithm}    & {Valid}     & {Loss}    & {Epochs} & {Batch} & {Learn rate} & {Decay} \\ \cmidrule(r){1-8} 
{MLP}                                               & {AdaDelta}     & {Train}     & {Entropy} & {5000}   & {16}    & {1.0}           & {0.0}   \\
{FCN}                                               & {Adam}         & {Train}     & {Entropy} & {2000}   & {16}    & {1}             & {0.0}   \\
{ResNet}                                            & {Adam}         & {Train}     & {Entropy} & {1500}   & {16}    & {1}             & {0.0}   \\
{Encoder}                                           & {Adam}         & {Train}     & {Entropy} & {100}    & {12}    & {1}             & {0.0}   \\
{MCNN}                                              & {Adam}         & {$Split_{20\%}$} & {Entropy} & {200}    & {256}   & {0.1}           & {0.0}   \\
{t-LeNet}                                           & {Adam}         & {Train}     & {Entropy} & {1000}   & {256}   & {0.01}          & {5}     \\
{MCDCNN}                                            & {SGD}          & {$Split_{33\%}$} & {Entropy} & {120}    & {16}    & {0.01}          & {5}     \\
{Time-CNN}                                          & {Adam}         & {Train}     & {Mse}     & {2000}   & {16}    & {1}             & {0.0}   \\
\bottomrule
\end{tabular}
\end{adjustbox}
\end{table}

Each experiment was conducted over five independent runs, encompassing the training, validation, and testing phases. The training set was used to fit the models, while the validation set served to determine the optimal predictive parameters. Finally, the test set was utilized to evaluate the performance of the models.

\subsection{Results and discussions}

As presented in Table~\ref{tab:results-ftir-degree}, the results demonstrate that GANet achieved its highest classification performance when degree centrality was employed as the network measure.

\begin{table}[htbp]
\centering
\caption[Results using \emph{degree} as the network measure for importance concept]{Results using \emph{degree} as the network measure for importance concept in terms of accuracy, sensitivity, specificity, and the harmonic mean of other columns for both pre-processing techniques investigated. The highest values are highlighted in bold.}
\label{tab:results-ftir-degree}
\begin{adjustbox}{width=\textwidth}
\begin{tabular}{l l c c c c}
\toprule
Metric & Config. & Accuracy & Sensitivity & Specificity & H. Mean \\
\midrule
\multirow{4}{*}{Cosine}
    & GANet-C & 0.65 & 0.63 & 0.70 & 0.65 \\
    & GANet-E & 0.68 & 0.51 & 0.83 & 0.64 \\
    & GANet-G & 0.65 & 0.63 & 0.70 & 0.65 \\
    & GANet-K & 0.63 & 0.59 & 0.70 & 0.63 \\
\midrule
\multirow{4}{*}{Euclidean}
    & GANet-C & 0.64 & \textbf{0.65} & 0.66 & 0.64 \\
    & GANet-E & \textbf{0.78} & 0.61 & \textbf{0.90} & \textbf{0.74} \\
    & GANet-G & 0.64 & \textbf{0.65} & 0.66 & 0.64 \\
    & GANet-K & 0.66 & \textbf{0.65} & 0.68 & 0.66 \\
\bottomrule
\end{tabular}
\end{adjustbox}
\end{table}

The results presented in Table~\ref{tab:results-ftir-aut} summarize the classification performance of various algorithms under different pre-processing strategies. The evaluation metrics include accuracy, sensitivity, specificity, and harmonic mean (H. Mean). Each algorithm was assessed using two pre-processing pipelines: (i) normalization by the amide peak, and (ii) smoothing with a Savitzky–Golay filter, differentiation and normalization by the amide peak.

The highest metric values for each algorithm under each pre-processing combination are highlighted in bold within the table.

Among the evaluated methods, GANet achieved the highest overall accuracy (0.78) and specificity (0.90), indicating strong performance in correctly identifying true negatives. Additionally, GANet attained the highest harmonic mean (0.74), reflecting a balanced trade-off between sensitivity and specificity compared to other methods.

In contrast, SVM exhibited lower performance across most metrics, with an accuracy of 0.56 and a harmonic mean of 0.37. Although it achieved a high specificity (0.88), its sensitivity was notably low (0.20), indicating a limited ability to identify positive instances. LDA outperformed SVM, achieving an accuracy of 0.61 and moderate sensitivity (0.50), but with a lower specificity (0.64) and a harmonic mean of 0.57.

FCN achieved a high specificity of 0.87, comparable to GANet’s performance; however, its sensitivity was extremely low (0.00), indicating a complete failure to detect true positive cases. This imbalance severely impacted its harmonic mean (0.00), suggesting that FCN may be overfitting to a single class within the dataset.

Time-CNN demonstrated a more balanced performance, with reasonable accuracy (0.65) and sensitivity (0.75), along with moderate specificity (0.33). This balance across metrics resulted in a harmonic mean of 0.47, indicating better overall robustness compared to many other algorithms evaluated.

Both MCNN and MLP exhibited a pattern of achieving perfect sensitivity (1.00), meaning they successfully identified all positive instances. However, they recorded specificity and harmonic mean values of 0.00, suggesting these models tended to classify nearly all samples as positive, leading to poor specificity and unbalanced overall performance.

TWIESN achieved one of the highest specificities (0.87), second only to GANet, along with a moderate accuracy of 0.72. However, its sensitivity remained low (0.20), resulting in a moderate harmonic mean of 0.54, reflecting its tendency to correctly identify negative cases while often missing positive instances.

For most algorithms, the \emph{Amide I} pre-processing method notably enhances specificity, leading to improved detection of true negatives. This effect is particularly evident in GANet, FCN, and TWIESN, which exhibit some of the highest specificity values in the results. By focusing on the Amide I region and truncating non-informative portions of the spectra, this method likely preserves the most discriminative spectral features, contributing to the relatively high accuracy and specificity observed. The truncation step appears to effectively reduce noise while retaining essential information, thereby creating a favorable performance balance for models that benefit from clear, well-defined feature boundaries, such as GANet.

Conversely, the \emph{Smoot., diff., norm.} pre-processing pipeline tends to improve sensitivity, which is advantageous when the primary objective is to maximize true positive detection. However, this improvement in sensitivity often comes at the expense of specificity, potentially due to over-smoothing or the inadvertent loss of fine-grained discriminatory features during aggressive pre-processing.

In summary, the proposed method, GANet, demonstrates the most balanced performance across accuracy, sensitivity, and specificity, as reflected in its high harmonic mean. While alternative methods such as MCNN and MLP achieve excellent sensitivity, they suffer from a lack of specificity, which undermines their overall reliability for balanced classification tasks. Consequently, GANet is better suited for applications that require minimizing both false positives and false negatives, offering a robust and reliable solution for complex classification challenges, including those involving ATR-FTIR saliva spectra in Autism Spectrum Disorder detection.

\begin{table}[htbp]
\centering
\caption[Classification results of low and high level techniques]{Classification results in terms of accuracy, sensitivity, specificity, and the harmonic mean of other columns. The best result from Table \ref{tab:results-ftir-degree} was used to compare with the low-level techniques as well as deep learning algorithms. The highest values for each type of pre-processing technique under each algorithm are in bold.}
\label{tab:results-ftir-aut}
\begin{adjustbox}{width=\textwidth}
\begin{tabular}{l l c c c c}
\toprule
Algorithm & Pre-processing method & Accuracy & Sensitivity & Specificity & H. Mean \\
\midrule
GANet (Proposed) & Amide I & \textbf{0.78} & 0.61 & \textbf{0.90} & \textbf{0.74} \\
SVM & Smoot., diff., norm. & 0.56 & 0.20 & 0.88 & 0.37 \\
LDA & Smoot., diff., norm. & 0.61 & 0.50 & 0.64 & 0.57 \\
FCN & Amide I & 0.63 & \textbf{0.87} & 0.0 & 0.0 \\
MCDCNN & Smoot., diff., norm. & 0.54 & 0.62 & 0.33 & 0.46 \\
MCNN & Amide I & 0.72 & 1.0 & 0.0 & 0.0 \\
MLP & Smoot., diff., norm. & 0.72 & 1.0 & 0.0 & 0.0 \\
ResNet & Smoot., diff., norm. & 0.45 & 0.50 & 0.33 & 0.41 \\
Time-CNN & Amide I & 0.72 & 0.75 & 0.66 & 0.71 \\
t-LeNet & Amide I & 0.72 & 1.0 & 0.0 & 0.0 \\
TWIESN & Amide I & 0.72 & \textbf{0.87} & 0.33 & 0.54 \\
\bottomrule
\end{tabular}
\end{adjustbox}
\end{table}

\subsection{SHapley Additive exPlanations (SHAP)}

SHAP explanations have become a leading approach for feature attribution in explainable artificial intelligence \cite{van2022tractability}. Based on game theory principles, SHAP quantifies the contribution of each individual feature to a model's prediction and provides insights into how features influence outputs. The summary plot can be interpreted as follows:

\begin{itemize}
    \item \emph{Vertical axis (feature waves)}: Each line corresponds to a feature, represented by its wave number from 900 cm$^{-1}$ to 1800 cm$^{-1}$. These are the most relevant features used in the model.
    \item \emph{Horizontal axis (SHAP value)}: Represents the contribution of a feature to the model's prediction for a specific instance. Positive SHAP values push the prediction toward the positive class, while negative SHAP values push it toward the negative class.
    \item \emph{Color gradient (feature value)}: Each dot's color represents the feature value, ranging from light orange (low values) to dark brown (high values). This allows visualization of how different feature values affect the prediction.
    \item \emph{Dot spread}: Each dot represents a single instance. The spread along the horizontal axis shows how a feature's influence varies across instances. A wide spread indicates variable impact, while a narrow spread suggests a more consistent effect.
\end{itemize}

\begin{figure}[htbp]
\centering
 \includegraphics[width=\textwidth]{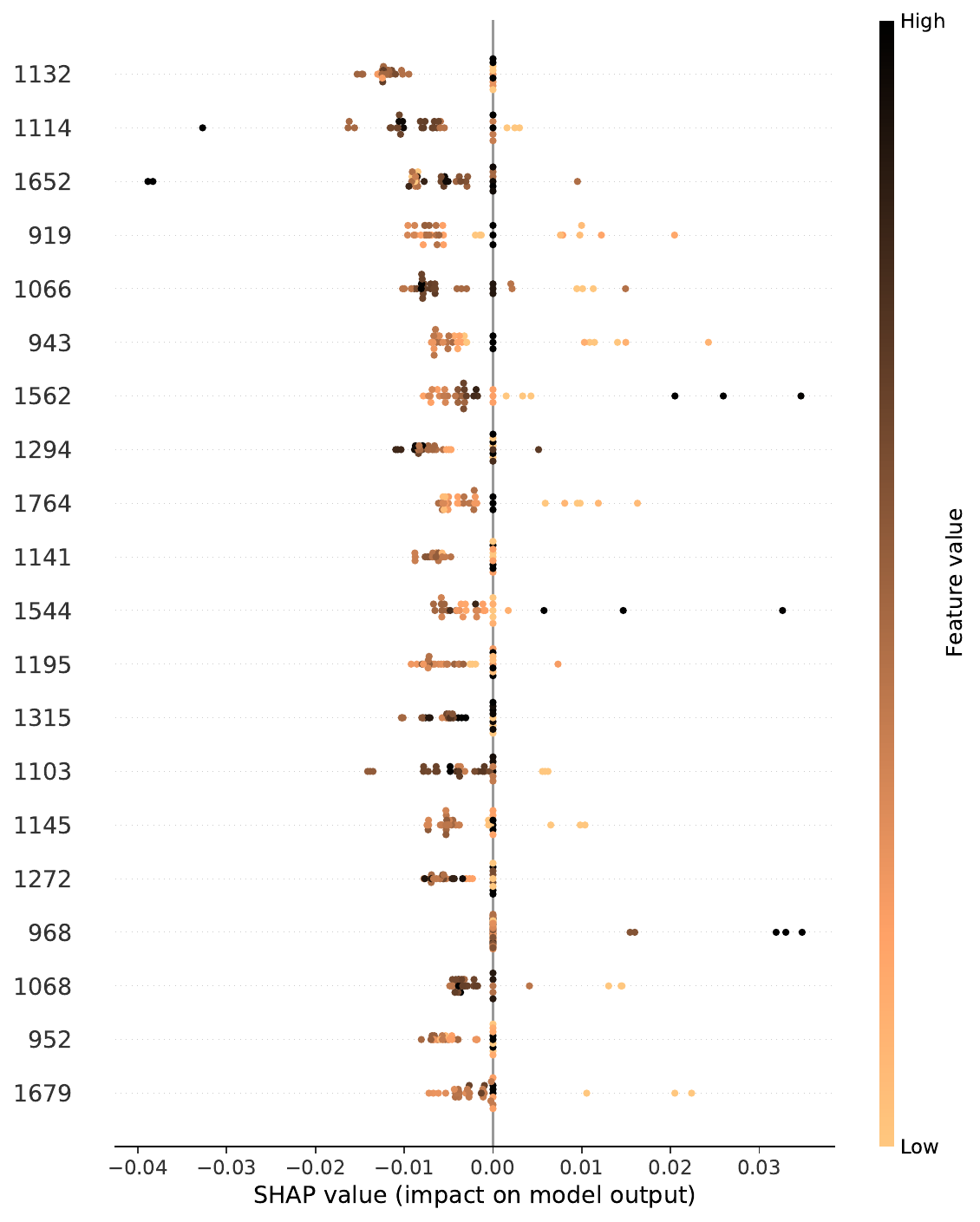}
\caption{SHAP summary plot insights into how each feature influences the GANet model.}
\label{fig:important-features}
\end{figure}

Fig. \ref{fig:important-features} shows the SHAP summary plot for GANet applied to the ASD dataset. Features with a wide spread of SHAP values, such as \emph{1132}, \emph{1114}, and \emph{1652}, exhibit variable influence on the predictions, either positively or negatively depending on their values. In contrast, features like \emph{1679} and \emph{952} have a narrow range of SHAP values around zero, indicating minimal influence and suggesting they could potentially be omitted without significantly affecting the performance of GANet.

Some features, such as \emph{1652} and \emph{1315}, tend to have mostly negative SHAP values, generally decreasing the predicted output. Others, like \emph{1114} and \emph{1764}, have both positive and negative values, indicating that their influence depends on the instance. 

For feature \emph{1132}, darker-colored dots (high values) are mostly on the positive side, suggesting that higher values increase the model output. Conversely, for feature \emph{1652}, lower values tend to decrease the prediction, as indicated by light-colored dots on the negative side. This helps identify features with monotonic effects versus those with more complex, non-linear relationships.

Features with high impact and large variability, such as \emph{1132} and \emph{1114}, are candidates for further investigation, as understanding their variable influence may reveal feature interactions or opportunities for model improvement.
 
\section{Conclusion} \label{sec:conclusion}

This study evaluated the effectiveness of the proposed GANet method in optimizing network structures for the classification of ATR-FTIR salivary spectra in ASD detection. Four GANet configurations, GANet-C, GANet-E, GANet-G, and GANet-K, were systematically compared against two widely used low-level classifiers, SVM and LDA. The results showed that GANet, when using the Amide I spectral region, and TWIESN with the Smoot., diff., norm. pre-processing pipeline, achieved high accuracy and balanced harmonic mean values across evaluation metrics. The Smoot., diff., norm. pre-processing consistently enhanced sensitivity, particularly for deep learning models such as Time-CNN and t-LeNet. In contrast, methods like SVM and FCN displayed biases toward specificity and sensitivity, respectively, highlighting potential overfitting or underfitting issues depending on the performance metric. These findings underscore the critical impact of pre-processing in balancing sensitivity, specificity, and overall performance in complex classification scenarios.

To facilitate high-level classification using importance-based characterization, a robust learning framework was developed, systematically exploring multiple genetic algorithm configurations to analyze their specific contributions and trade-offs in classification accuracy, network efficiency, and computational cost. This structured approach enabled the effective optimization of classification workflows within complex network structures.

The proposed GANet was rigorously evaluated against established classification approaches using FTIR-processed salivary data. Results demonstrated that GANet effectively captured and leveraged complex relationships within the data, establishing it as a viable and competitive alternative for high-level classification tasks utilizing network-based methodologies. Overall, this research introduced an innovative bio-inspired approach to network construction and optimization for high-level classification within complex networks, laying a strong foundation for future studies that integrate biological and spectral data to enhance accuracy and efficiency in data analysis. This is particularly relevant for advancing ASD detection using non-invasive salivary biomarkers, supporting the development of sustainable and precise diagnostic tools in healthcare.

Nevertheless, some limitations must be acknowledged. The study uses 159 spectra from 53 participants. While carefully balanced between ASD and controls, the dataset remains relatively small for machine learning, which may limit generalizability to broader populations. GANet achieved high specificity (0.90) but only moderate sensitivity (0.61), indicating that while the model reliably identifies non-ASD cases, it is less effective at detecting ASD cases, which could result in false negatives in real diagnostic settings. Additionally, replicates (triplicate spectra per participant) contribute to stability but do not increase biological diversity, highlighting the need for larger and more heterogeneous datasets.

Future work should focus on validating GANet on larger, more diverse populations and across multiple collection sites to confirm its robustness and generalizability. The generation and use of synthetic spectral data could also help to address sample size limitations and improve model stability. Furthermore, investigating parallelization strategies or surrogate-assisted evolutionary optimization could reduce the computational cost of GA training, making GANet more scalable and accessible for real-world applications.



\end{document}